\documentclass[conference]{IEEEtran}
\usepackage[pdftex]{graphicx}
\usepackage{amsmath}
\usepackage{algorithmic}
\usepackage{array}
\usepackage{stfloats}
\usepackage{url}
\usepackage{amsmath}
\usepackage[linesnumbered,ruled]{algorithm2e}
\usepackage{multirow}

\usepackage[caption=false,font=footnotesize]{subfig}

\begin{document}
\title{Evolving Deep Convolutional Neural Networks by Variable-length Particle Swarm Optimization for Image Classification}

\author{
\IEEEauthorblockN{Bin Wang, Yanan Sun, Bing Xue and Mengjie Zhang}
\IEEEauthorblockA{School of Engineering and Computer Science\\
	Victoria University of Wellington,
	PO Box 600, Wellington 6140, NEW ZEALAND}
\IEEEauthorblockA{Emails: wangbin@myvuw.ac.nz, \{yanan.sun, bing.xue, mengjie.zhang\}@ecs.vuw.ac.nz}
}


%Telephone: (800) 555--1212, Fax: (888) 555--1212}


\maketitle

\begin{abstract}
Convolutional neural networks (CNNs) are one of the most effective deep learning methods to solve image classification problems, but the best architecture of a CNN to solve a specific problem can be extremely complicated and hard to design. This paper focuses on utilising Particle Swarm Optimisation (PSO) to automatically search for the optimal architecture of CNNs without any manual work involved. In order to achieve the goal, three improvements are made based on traditional PSO. 
%First, a novel Internet Protocol based (IP-based) encoding strategy of using network IP structure to encode CNN layers into particle vectors is proposed and the IP-based PSO method will be called IPPSO in this paper;% 
First, a novel encoding strategy inspired by computer networks which empowers particle vectors to easily encode CNN layers is proposed; Second, in order to allow the proposed method to learn variable-length CNN architectures, a Disabled layer is designed to hide some dimensions of the particle vector to achieve variable-length particles; Third, since the learning process on large data is slow, partial datasets are randomly picked for the evaluation to dramatically speed it up. The proposed algorithm is examined and compared with 12 existing algorithms including the state-of-art methods on three widely used image classification benchmark datasets. The experimental results show that the proposed algorithm is a strong competitor to the state-of-art algorithms in terms of classification error. This is the first work using PSO for automatically evolving the architectures of CNNs.
\end{abstract}


\ifCLASSOPTIONpeerreview
\begin{center} \bfseries EDICS Category: 3-BBND \end{center}
\fi
\IEEEpeerreviewmaketitle

\section{Introduction}
Convolutional neural networks (CNNs) have demonstrated exceptional superiority in numerous machine learning tasks, such as speech recognition \cite{CNNspeech:Ossama}, sentence classification \cite{CNNsentence:Yoon} and image classification \cite{ImageNet:Alex}. However, designing the architectures of CNNs for specific tasks can be extremely complex, which can be seen from some existing efforts done by researchers, such as LeNet \cite{ZipcodeRecognition:LeCun}\cite{DocumentRecognition:LeCun}, AlexNet \cite{ImageNet:Alex}, VGGNet \cite{CNNverydeep:Simonyan} and GoogLeNet \cite{CNNdeeper:Szegedy}. %When the architecture of CNN gets deeper, the hyper-parameters and weights become more complex which makes the further improvement of the CNN architecture harder.
In addition, one cannot expect to get the optimal performance by applying the same architecture on various tasks, and the CNN architecture needs to be adjusted for each specific task, which will bring tremendous work as there are a large number of types of machine learning tasks in industry.

In order to solve the complex problem of the CNN architecture design, evolutionary computation (EC) has recently been leveraged to automatically design the architecture without any human effort involved. Interested researchers have done excellent work on the automatic design of the CNN architectures by using Genetic Programming (GP) \cite{CNNGP:Suganuma} and Genetic Algorithms (GAs) \cite{CNNevolve:Stanley}, such as Large-scale evolution of image classifiers (LEIC) method \cite{LEIC:Real} recently proposed by Google, which have shown that EC can be used in learning CNN architectures that are competitive with the state-of-art algorithms designed by humans. However, the learning process for large data is too slow due to the high computational cost for most of the methods and it might not be practical for industrial use.

A lot of work has been done in order to improve using EC to evolve a CNN architecture, such as the recent proposed EvoCNN using GAs \cite{EvolveCNN:Yanan}. One of the improvements in EvoCNN is that during the fitness evaluation, instead of training the model for 25,600 steps in LEIC, it only trains each individual by 10 epochs, which dramatically speeds up the learning process. The rationale behind EvoCNN using 10 epochs is that the researchers believe that training 10 epochs can obtain the major trend of the CNN architecture, which would be decisive to the final performance of a model, having been verified by their experiments. 

%However, not a lot of research has been done by using other Evolutionary Computation methods to evolve the architectures of CNNs. In this paper, we would like to explore using PSO to evolve the architectures of CNNs for image classification as PSO has advantages of easy implementation and lower computational cost. First of all, We will develop a variable-length PSO with a novel IP-based encoding strategy to cope with variable-length CNN architectures. In addition, we will adjust the fitness evaluation method proposed in EvoCNN by using partial dataset \footnote{we use 10\% to 20\% of the whole training set for fitness evaluations depending on the specific problem.} instead of the whole training set for fitness evaluation as training only 10\% to 20\% of the dataset can significantly speed up the training process which then saves tremendous time of the whole evolution process and we believe using partial dataset with 10 epochs is able to learn the trend of a CNN architecture which will be confirmed in the experiments. Overall, the proposed IPPSO method is able to drastically improve the speed of the learning process along with obtaining a very competitive classification performance. 

However, not a lot of research has been done by using other EC methods to evolve the architectures of CNNs, so we would like to explore some other major EC methods for evolving the architectures of CNNs without any human interference. Since Particle Swarm Optimisation (PSO) has the advantages of easy implementation, lower computational cost, and fewer parameters to adjust, and it has never been utilised to evolve the architectures of CNNs, PSO is chosen in this paper. However, the fix-length encoding of the particle in traditional PSO is a big challenge for evolving the architectures of CNN as the optimal CNN architecture varies for different tasks, so a novel flexible encoding scheme is proposed to break the fix-length constraint, which would be the most fundamental part of the proposed algorithm in this paper.

\subsection{Goal}
The overall goal of this paper is to design and develop an effective and efficient PSO method to automatically discover good architectures of CNNs. The specific objectives of this paper are to

\begin{enumerate}
%	\item Design a new particle encoding scheme, which has the ability of effectively encoding a CNN layer of different types into a network interface \footnote{Please refer to Section \ref{sec:IPAddress} if Network Interface does not sound familiar}, which is stored in order in the particle vector to represent a CNN architecture, and develop a new PSO algorithm based on the novel encoding strategy. 
	\item Design a new particle encoding scheme that has the ability of effectively encoding a CNN architecture, and develop a new PSO algorithm based on the novel encoding strategy. 
%	\item Design the IP subnets especially a disabled subnet to enable the IPPSO method to learn variable-length CNN architectures as the optimal architectures of CNNs for different tasks could vary in length. 
%	\item Design a method to break the constraint of the fix-length encoding of traditional PSO in order to learn variable-length architectures of CNNs. To be specific, along with the other CNN layers, we introduce a new layer called Disabled layer, which is detailed in Section \ref{sec:ParticleEncodingStrategy}. The network interface representing the Disabled layer would be hidden when decoding the particle, which attains a variable-length particle.
	\item Design a method to break the constraint of the fix-length encoding of traditional PSO in order to learn variable-length architectures of CNNs. We will introduce a new layer called Disabled layer to attain a variable-length particle.
	\item Propose a fitness evaluation method using a partial dataset instead of the whole dataset to significantly speed up the evolutionary process. 
\end{enumerate} 

%\subsection{Organisation}
%The remaining parts of this paper are organised as follows: first of all, the background of the CNN, PSO and IP address are
%reviewed in Section \ref{sec:Background}. In addition, the framework and details of each step in the proposed algorithm are elaborated in Section \ref{sec:ProposedAlgorithm}. Furthermore, the experiment design and experimental results of the proposed algorithm are shown in Sections \ref{sec:EPDesign} and \ref{sec:EPResults}, respectively. Last but not least, the conclusions and future work are discussed in Section \ref{sec:Conclusion}. 
%Next, further discussions are made in Section \ref{sec:FurtherDiscussion}.

\section{Background}\label{sec:Background}

\subsection{CNN architecture}\label{sec:CNNArchitecture}

\begin{figure}[!t]
	\centering
	\includegraphics[width=3.5in]{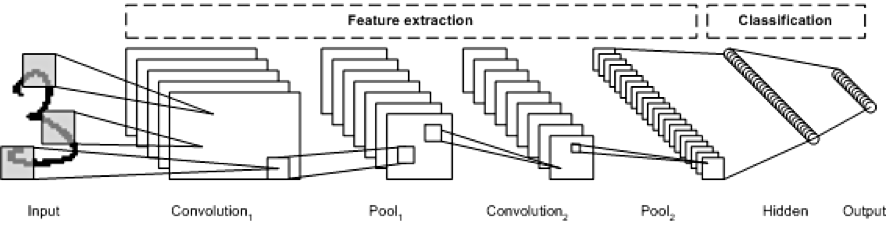}
	\caption{An general architecture of the Convolutional Neural Network \cite{CNNFig:Tim}}
	\label{fig:CNNArchitecture}
\end{figure}

Fig. \ref{fig:CNNArchitecture} exhibits a general architecture of a CNN with two convolutional (Conv) layers, two pooling layers, and two Fully-connected layers - one hidden layer and one output layer at the end \cite{CNNFig:Tim}. It is well-known that when designing a deep CNN architecture, the number of Conv layers, Pooling layers and Fully-connected layers before the last output layer have to be properly defined along with their positions and configurations. In terms of the configuration, apart from the output layer which will almost be a number of neurons having the number of classes in the classification problem as the size, different types of layers have different configurations as follows: Filter size, stride size and feature maps are the main attributes of the configuration for the Conv layer; Kernel size, stride size and pooling type - max-pooling or average-pooling, are the important parameters for the configuration of the Pooling layer; and the number of neurons is the key attribute of the Fully-connected layer. %After the layers have been designed, a proper weight initialisation method has to be chosen, and Xavier weight initialisation \cite{WeightIniti:Glorot} will be chosen as it has been proved as an effective way and has been implemented in most of Deep Learning frameworks.

\subsection{Particle Swarm Optimisation}

Particle Swarm Optimization (PSO) is a population-based algorithm, motivated by the social behaviour of fish schooling or bird flocking \cite{PSOIntro:Kennedy} \cite{PSOIntro:Eberhart}, commonly used for solving optimization problems without rich domain knowledge \cite{PSOIntro:Yanan}. In PSO, the population is composed of a certain number of particles each of which represents a solution, and particles fly in the search space to find the best solution by updating velocity and particle vector according to Equations (\ref{eq:UpdateV}) and (\ref{eq:UpdateX}), respectively, where $v_{id}$ represents the velocity of the particle $i$ in the $d$th dimension, $x_{id}$ represents the position of particle $i$ in the $d$th dimension, $P_{id}$ and $P_{gd}$ are the local best and the global best in the $d$th dimension, $r_{1}, r_{2}$ are random numbers between 0 and 1, $w, c_{1}$ and $c_{2}$ are PSO parameters used to tweak the performance. 

\begin{equation}\label{eq:UpdateV}
	\begin{aligned}
	v_{id}(t+1) = w * v_{id}(t) + c_{1} * r_{1} * (P_{id} - x_{id}(t)) + \\
	c_{2} * r_{2} * (P_{gd} - x_{id}(t))
	\end{aligned}
\end{equation}

\begin{equation}\label{eq:UpdateX}
	x_{id}(t+1) = x_{id}(t) + v_{id}(t+1)
\end{equation}

\subsection{Internet Protocol address}\label{sec:IPAddress}

An Internet Protocol address (IP address) is a numerical label assigned to each device connected to a computer network that uses the Internet Protocol for communication \cite{IP:Postel}. In order to identify the network of an IP address, the subnet is introduced, which is often described in CIDR (Classless Inter-Domain Routing) style \cite{CIDR:Fuller} by combining the starting IP address and the length of the subnet mask together. Both the IP address used for identifying the host and its corresponding subnet used for distinguishing different networks are carried by a network interface. For example, a standard IP address of a 32-bit number could be 192.168.1.251, and a standard subnet carries the starting IP address and the length of the subnet mask could be 192.168.1.0/8, which indicates the IP address in the subnet starts from 192.168.1.0 and the length of the subnet mask is 8 defining an IP range from 192.168.1.0 to 192.168.1.255.

Although the outlook of the IP address is a sequence of decimal numbers delimited by full stops, the binary string under the hood is actually used for the network identification, which inspires the new PSO encoding scheme. As there are several attributes in the configuration of each type of CNN layers, each of which is an integer value within a range, each value of the attribute can be smoothly converted to a binary string, and several binary strings, each of which represents the value of an attribute, can be concatenated to a large binary string to represent the whole configuration of a specific layer. It is obvious that the binary string suits the requirement of encoding CNN layers to particles. However, in this way, a huge number converted from the binary string will have to be used as one dimension of the particle vector, which may result in a horrendous searching time in PSO. On the other hand, in the IP structure, instead of utilising one huge integer to mark the identification (ID) of a device in a large network, in order to make the IP address readable and memorable, it divides a huge ID number into several decimal values less than 256, each of which is stored in one byte of the IP address. In this way, the binary string can be divided into several bytes, and each byte comprises one dimension of the particle vector. The convergence of PSO can be facilitated by splitting one dimension of a large number to several dimensions of small numbers because in each round of the particle updates, all of the dimensions can be concurrently learned and the search space of one split dimension is much smaller. In this paper, the new particle encoding scheme will use this idea to gain the flexibility of encoding various types of layers into a particle, and drastically cut down the learning process, which will be described in the next section.

%\subsection{PSO related work in Deep CNN}
%
%As it is shown in Section \ref{sec:CNNArchitecture}, there are a lot of factors that need to be considered when doing the CNN architecture design and it is obviously an extremely hard work for humans to manually achieve the best design. As a result, researchers have been trying to leverage Evolutionary Algorithm to evolve CNN architectures such as LEIC method \cite{LEIC:Real}; However, the learning process is too slow. In this paper, we will design a more efficient PSO method. Since the length of the best CNN architecture on specific problems is not fixed while the particle length of traditional PSO is fixed, the challenge of the proposed PSO method will be inventing a new encoding scheme to cope with the variable-length problem. 

\section{The proposed algorithm}\label{sec:ProposedAlgorithm}
In this section, the new IP based PSO (IPPSO) method for evolving deep CNNs will be presented in detail.

\subsection{Algorithm Overview}
\begin{algorithm}
	\caption{Framework of IPPSO}
	\label{alg:framework}
	\begin{algorithmic}
		\renewcommand{\algorithmicrequire}{\textbf{Input:}}
		\renewcommand{\algorithmicensure}{\textbf{Output:}}
		\STATE $P \leftarrow$ Initialize the population with the proposed particle encoding strategy;
		\STATE $P_{id} \leftarrow empty$;
		\STATE $P_{gd} \leftarrow empty$;
		\WHILE{termination criterion is not satisﬁed}
			\STATE update velocity and position of each particle shown in Algorithm \ref{alg:update};
			\STATE evaluate the fitness value of each particle;
			\STATE update $P_{id}$ and $P_{gd}$;
		\ENDWHILE		
	\end{algorithmic}
\end{algorithm}

Algorithm \ref{alg:framework} outlines the framework of the proposed algorithm. There are mainly three steps to initialise the population by using the particle encoding strategy which will be described in Section \ref{sec:ParticleEncodingStrategy}, to update the position and velocity, and to check whether the termination criterion meets.

\subsection{Particle Encoding Strategy}\label{sec:ParticleEncodingStrategy}

The IPPSO encoding strategy is inspired by how the Network IP address works. Although the CNN architecture is comprised of three types of layers - Convolutional Layer, Pooling Layer, and Fully-Connected Layer, and the encoded information of different types of layers varies in terms of both the number of parameters and the range in each parameter shown in Table \ref{table:CNNFields}, a Network IP address with a fixed length of enough capacity can be designed to accommodate all the types of CNN layers, and then the Network IP can be divided into numerous subsets, each of which can be used to define a specific type of CNN layers.

First of all, the length of the binary string under the IP-based encoding scheme needs to be designed. With regard to Conv layers, firstly, there are three key parameters - filter size, number of feature maps and stride size listed in the column of Parameter in Table \ref{table:CNNFields}, which are the fundamental factors affecting the performance of CNNs; Secondly, based on the size of benchmark datasets, the range of the parameters are set to [1,8], [1,128] and [1,4] for the aforementioned three parameters, respectively, shown in the column of Range in table \ref{table:CNNFields}; Thirdly, taking a CNN architecture with the filter size of 2, number of feature maps of 7 and stride size of 2 as an example, the decimal values can be converted to the binary strings of 001, 000 1111 and 01, where the binary string converted from the decimal value is filled with 0s until the length reaches the corresponding number of bits, illustrated in the column of Example Value in Table \ref{table:CNNFields}. Lastly, the total number of bits of 12 and the sample binary string of 001 000 1111 01 by concatenating the binary strings of the three parameters are displayed in the summary row of Conv layer in Table \ref{table:CNNFields}. In terms of Pooling layers and Fully-connected layers, the total number of bits and the sample binary string can be obtained by following the same process of Conv layers, which are listed in the summary rows of Pooling and Fully-connected layers in Table \ref{table:CNNFields}.  As the largest number of bits to represent a layer is 12 as shown in Table \ref{table:CNNFields} and the unit of an IP address is one byte - 8 bits, there will be 2 bytes required to accommodate the 12 bits IP address. 

\begin{table}[!t]
	%% increase table row spacing, adjust to taste
	\renewcommand{\arraystretch}{1.3}
	% if using array.sty, it might be a good idea to tweak the value of
	% \extrarowheight as needed to properly center the text within the cells
	\caption{The parameters of different types of CNN layers - Convolutional, Pooling, Fully-connected and Disabled layer with an example in the Example column}
	\label{table:CNNFields}
	\centering
	%% Some packages, such as MDW tools, offer better commands for making tables
	%% than the plain LaTeX2e tabular which is used here.
	\begin{tabular}{|p{1.5cm}|p{1.5cm}|p{1cm}|p{0.5cm}|p{2cm}|}
		\hline
		Layer Type & Parameter & Range & \# of Bits & Example Value\\
		\hline
		Conv & Filter size & [1,8] & 3 & 2(001)\\
		\hline
		& \# of feature maps & [1,128] & 7 & 32(000 1111)\\
		\hline
		& Stride size & [1,4] & 2 & 2(01)\\
		\hline
		& \textbf{Summary} &  & 12 & 001 000 1111 01\\
		\hline
		Pooling & Kernel size & [1,4] & 2 & 2(01)\\
		\hline
		& Stride size & [1,4] & 2 & 2(01)\\
		\hline
		& Type: 1(maximal), 2(average) & [1,2] & 1 & 2(1)\\
		\hline
		& Place holder & [1,128] & 6 & 32(00 1111)\\
		\hline
		& \textbf{Summary} &  & 11 & 01 01 0 00 1111\\
		\hline
		Fully-connected & \# of Neurons & [1,2048] & 11 & 1024(011 11111111)\\
		\hline
		& \textbf{Summary} &  & 11 & 011 11111111\\
		\hline
		Disabled & Place holder & [1,2048] & 11 & 1024(011 11111111)\\
		\hline
		& \textbf{Summary} &  & 11 & 011 11111111\\
		\hline
	\end{tabular}
\end{table}

In addition, the subnets for all types of CNN layers need to be defined according to the number of bits of each layer illustrated in Table \ref{table:CNNFields} and CIDR style will be used to represent the subnet. As there are three types of CNN layers, we need to define three subnets with enough capacity to represent all the types of layers. Starting with the Conv layer, 0.0 is designed as the starting IP address of the subnet; in addition, the total length of the designed 2-byte IP address is 16 and the total number of bits required by the Conv layer is 12, so the subnet mask length is 4 calculated by subtracting the total number of bits from the length of the IP address, which brings the subnet representation to 0.0/4 with the range from 0.0 to 15.255. Regarding the Pooling layer, the starting IP address is 16.0 obtained by adding 1 to the last IP address of the Conv layer, and the subnet mask length is 5 calculated in the same way as that of the Conv layer, which results in 16.0/5 with the range from 16.0 to 23.255 as the subnet representation of the Pooling layer. Similarly, the subnet 24.0/5 with the range from 24.0 to 31.255 is designed as the subnet of the Fully-connected layer. In order to make the subnets clear, all of the subnets are depicted in Table \ref{table:Subnets}.

As the particle length of PSO is fixed after initialisation, in order to cope with the variable-length of the architectures of CNNs, an effective way of disabling some of the layers in the encoded particle vector will be used to achieve this purpose. Therefore, another layer type called the Disabled layer and the corresponding subnet named the Disabled subnet are introduced. To achieve a comparable probability for the Disabled layer, the least total number of bits of 11 among all three types of CNN layers is set as the number of bits of the Disabled layer, so the disabled subnet comes to 32.0/5 with the range from 32.0 to 39.255, shown in Table \ref{table:Subnets}, where each layer will be encoded into an IP address of 2 bytes. Table \ref{table:IPExample} shows how the example in Table \ref{table:CNNFields} is encoded into IP addresses by combining all the binary string of each parameter of a specific layer into one binary string, filling the combined binary string with zeros until reaching the length of 2 bytes, applying the subnet mask on the binary string, and converting the final binary string to an IP address with one byte as a unit delimited by full stops. For instance, the sample binary string of the Conv layer in Table \ref{table:CNNFields} is 001 000 1111 01, which is filled to 0000 001 000 1111 01 to reach the length of 2 bytes; then 2-byte binary string - 0000 0010 and 0011 1101, can be obtained by applying the subnet mask, in which the starting IP address of the subnet is added to the binary string; Finally, the IP address of 2.61 is achieved by converting the first byte to the decimal value of 2 and the second byte to 61.

\begin{table}[!t]
	%% increase table row spacing, adjust to taste
	\renewcommand{\arraystretch}{1.3}
	% if using array.sty, it might be a good idea to tweak the value of
	% \extrarowheight as needed to properly center the text within the cells
	\caption{Four subnets distributed to the three types of CNN layers and the disabled layer}
	\label{table:Subnets}
	\centering
	%% Some packages, such as MDW tools, offer better commands for making tables
	%% than the plain LaTeX2e tabular which is used here.
	\begin{tabular}{|c|c|c|}
		\hline
		Layer type & Subnet(CIDR) & IP Range\\
		\hline
		Convolutional Layer & 0.0/4 & 0.0-15.255\\
		\hline
		Fully-Connected Layer & 16.0/5 & 16.0-23.255\\
		\hline
		Pooling Layer & 24.0/5 & 24.0-31.255\\
		\hline
		Disabled Layer & 32.0/5 & 32.0-39.255\\
		\hline
	\end{tabular}
\end{table}

\begin{table}[!t]
	%% increase table row spacing, adjust to taste
	\renewcommand{\arraystretch}{1.3}
	% if using array.sty, it might be a good idea to tweak the value of
	% \extrarowheight as needed to properly center the text within the cells
	\caption{An example of IP addresses - one for each type of CNN layers}
	\label{table:IPExample}
	\centering
	%% Some packages, such as MDW tools, offer better commands for making tables
	%% than the plain LaTeX2e tabular which is used here.
	\begin{tabular}{|c|c|c|}
		\hline
		Layer type & Binary (filled to 2 bytes) & IP address\\
		\hline
		Convolutional Layer & (0000)001 000 1111 01 & 2.61\\
		\hline
		Pooling Layer & (00000)01 01 0 00 1111 & 18.143\\
		\hline
		Fully-Connected Layer & (00000)011 11111111 & 27.255\\
		\hline
		Disabled Layer & (00000)01111111111 & 35.255\\
		\hline
	\end{tabular}
\end{table}

After converting each layer into a 2-byte IP address, the position and velocity of PSO can be defined. However, there are a few parameters that need to be mentioned first - max\_length(maximum number of CNN layers), max\_fully\_connected(maximum Fully-connected layers with the constraint of at least one Fully-connected layer) listed in Table \ref{table:ParameterList}. The encoded data type of the position and the velocity will be a byte array with a fixed length of maximum\_length * 2 and each byte will be deemed as one dimension of the particle.

\begin{figure}[!t]
	\centering
	\includegraphics[width=3.5in]{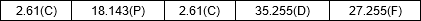}
	\caption{An example of IP addresses in a particle containing 5 CNN layers}
	\label{fig:ParticleIP}
\end{figure}

\begin{figure}[!t]
	\centering
	\includegraphics[width=3.5in]{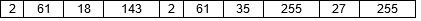}
	\caption{An example of a particle vector with 5 CNN layers encoded}
	\label{fig:ParticleVector}
\end{figure}

Here is an example of a particle vector to explain how the CNN architecture is encoded and how it copes with variable-length of CNN architecture. Assume max\_length is 5, a sequence of IP addresses representing a CNN architecture with the maximum number of 5 layers can be encoded into 5 IP addresses in Fig. \ref{fig:ParticleIP} by using the sample IP addresses in Table \ref{table:IPExample}, where C represents a Conv layer, P represents a Pooling layer, F represents a Fully-connected layer, and D represents a Disabled layer. The corresponding particle vector with the dimension of 10 is shown in Fig. \ref{fig:ParticleVector}. Since there is one Disabled layer in the example, the actual number of layers is 4. However, after a few PSO updates, the seventh dimension and the eighth dimension of the particle vector may become 18 and 143, respectively, which turns the third IP address representing a Disabled layers to a Pooling layer, so the updated particle carries a CNN architecture of 5 layers; Conversely, after a few updates, the fifth dimension and the sixth dimension of the particle vector may become 35 and 255, respectively, which makes the third IP address fall into the disabled subnet, so the actual number of layers is 3. To conclude, as shown in this example, the particle with IPPSO encoding scheme is capable of representing variable-length architectures of CNNs - 3, 4 and 5 in this example.

\subsection{Population Initialisation}

In terms of the population initialisation, after the size of the population is set up, individuals are randomly created until reaching the population size. 
For each individual, an empty vector is initialised first, and each element in it will be used to store a Network Interface containing the IP address and subnet information. The first element will always be a Conv layer; From the second to $(max\_length-max\_fully\_connected)$ layer, each element can be filled with a Conv layer, Pooling layer or Disabled layer; From $(max\_length-max\_fully\_connected)$ to $(max\_length-1)$ layer, it can be filled with any of the four types of layers until the first Fully-connected is added, and after that only Fully-connected layers or Disabled layers are allowed; The last element will always be a Fully-connected layer with the size the same as the number of classes. In addition, each layer will be generated with the random settings - a random IP address in a valid subnet.

\subsection{Fitness Evaluation}
\begin{algorithm}
	\caption{Fitness Evaluation}
	\label{alg:fitness}
	\begin{algorithmic}
		\renewcommand{\algorithmicrequire}{\textbf{Input:}}
		\renewcommand{\algorithmicensure}{\textbf{Output:}}
		\REQUIRE The population $P$, the training epoch number $k$, the training set $D_{train}$, the fitness evaluation dataset $D_{fitness}$, the batch size $batch\_size$;
		\ENSURE The population with fitness $P$;
		\FOR{individual $s \textbf{ in } P$}
			\STATE $i \leftarrow 1$;
			\WHILE{$i<=k$}
				\STATE $\textit{Train the connection weights of the CNN } \newline \textit{represented by individual s}$;
			\ENDWHILE
			\STATE $accy\_list \leftarrow$ Batch-evaluate the trained model on the dataset $D_{fitness}$ with the batch size $batch\_size$ and store the accuracy for each batch;
%			\STATE $(mean, stddev) \leftarrow$ Calculate the mean value and standard deviation of  $acc\_list$;
%			\STATE $num\_of\_connections \leftarrow$ Calculate the number of connections in s;
%			\STATE $fitness \leftarrow (mean, -stddev, -num\_of\_connections)$;
			\STATE $mean \leftarrow$ Calculate the mean value of  $acc\_list$
			\STATE $fitness \leftarrow mean$;
			\STATE $P \leftarrow$ Update the fitness of the individual $ind$ in the population $P$;
		\ENDFOR	
		\RETURN $P$	
	\end{algorithmic}
\end{algorithm}

Before performing the fitness evaluation, a proper weight initialisation method has to be chosen, and Xavier weight initialisation \cite{WeightIniti:Glorot} is chosen as it has been proved as an effective way, and has been implemented in most of Deep Learning frameworks. With regard to the fitness evaluation (shown in Algorithm \ref{alg:fitness}), each individual is decoded to a CNN architecture with its settings, which will be trained for $k$ epochs on the first part of the training dataset. Then the partially trained CNN will be batch-evaluated on the second part of the training dataset, which will produce a series of accuracies. Finally, we calculate the mean value of the accuracies for each individual, which will be stored as the individual fitness. 
%we calculate the mean and standard deviation of the accuracies for each individual, which will be stored as the individual fitness along with the number of connections in a CNN. 
%For comparing the fitness of individuals, the mean value, standard deviation and the number of parameters will be used in the comparison, i.e. compare mean value first, if mean value is equal, compare standard deviation, and if standard deviation is the same, compare the number of parameters.

\subsection{Update Particle with Velocity Clamping}
\begin{algorithm}
	\caption{Update Particle with Velocity Clamping}
	\label{alg:update}
	\begin{algorithmic}
		\renewcommand{\algorithmicrequire}{\textbf{Input:}}
		\renewcommand{\algorithmicensure}{\textbf{Output:}}
		\REQUIRE particle individual vector $ind$, acceleration coefficient array for $P_{id}$ $c_{1}$, acceleration coefficient array for $P_{gd}$ $c_{2}$, inertia weight $w$, max velocity array $v_{max}$;
		\ENSURE updated individual vector $ind$;
		\FOR{element $interface \textbf{ in } ind$}
			\STATE $i \leftarrow 0$;
			\FOR{$i <$ number of bytes of IP address in $interface$}
				\STATE $x \leftarrow$ the $i$th byte of the IP address in the $interface$;
				\STATE $(r_{1}, r_{2})) \leftarrow$ uniformly generate $r_{1}, r_{2}$ between [0, 1];
				\STATE $v_{new} \leftarrow$ Update velocity based on Equation \ref{eq:UpdateVNew};
				\STATE $v_{new} \leftarrow$ Apply velocity clamping using $v_{max}$;
				\STATE $x_{new} \leftarrow x + v_{new}$
				\IF{$x_{new} > 255$}
					\STATE $x_{new} \leftarrow x_{new}-255$;
				\ENDIF
			\ENDFOR
		\ENDFOR
		\STATE $fitness \leftarrow$ evaluate the updated individual $ind$;
		\STATE $(P_{id}, P_{gd}) \leftarrow$ Update $pbest$ and $gbest$ by comparing their $fitness$;
		\RETURN $ind$
	\end{algorithmic}
\end{algorithm}

In Algorithm \ref{alg:update}, as each layer is encoded into an interface with 2 bytes in the particle vector, and we want to control the acceleration coefficients for each byte, the two acceleration coefficients implemented as two float arrays with the size of 2 are required shown in Equation \ref{eq:UpdateVNew}. $v$ and $x$ are decimal values of the $i$th byte of the 2-byte IP address and its corresponding velocity, $P_{id}$ and $P_{gd}$ are decimal values of the $i$th byte of the IP address of the local best and global bet, respectively, and $w, r_{1}, r_{2}$ are the same as traditional PSO in Equation \ref{eq:UpdateV}. The major difference is how the acceleration coefficients are implemented - $c_{1}[i]$ and $c_{2}[i]$ are the acceleration coefficients for the $i$th byte of the IP address, where $i$ is 1 or 2 in the case of 2-byte IP encoding, comparing to a singular value for each of the acceleration coefficients in traditional PSO. The reason of separating the acceleration coefficients for each byte of the IP address is that different parameters may fall into different bytes of the IP address, and the ability to explore a specific parameter more than others may be needed when fine-tuning the learning process.

After the coefficients defined, we go through each byte in the particle and update the velocity and position by using the corresponding coefficients for that byte. Since there are some constraints for each interface in the particle vector according to its position in the particle vector, e.g. the second interface can only be a Conv layer, Pooling layer or Disabled layer, the new interface needs to be replaced by an interface with a random IP address in a valid subnet if the new interface does not fall in a valid subnet. After all the bytes being updated, the new particle is evaluated, and the fitness value is compared with the local best and global best in order to update the two bests if needed.

\begin{equation}\label{eq:UpdateVNew}
\begin{aligned}
v_{new} = w * v + c_{1}[i] * r_{1} * (P_{id} - x) + c_{2}[i] * r_{2} * (P_{gd} - x)
\end{aligned}
\end{equation}

\subsection{Best Individual Selection and Decoding}

The global best of PSO will be reported as the best individual. In terms of the decoding, a list of network interfaces - stored in every 2 bytes from left to right in the particle vector of the global best, can be extracted from a particle vector. According to the subnets in Table \ref{table:Subnets} the type of layer can be distinguished, and then based on Table \ref{table:CNNFields} the IP address can be decoded into different sets of binary string, which indicate the parameter values of the layer. After decoding all the interfaces in the global best, the final CNN architecture can be attained by connecting all of the decoded layers in the same order as that of the interfaces in the particle vector.

\section{Experiment design}\label{sec:EPDesign}

%In order to examine the performance of the proposed IPPSO for evolving CNN, a series of experiments is designed and performed on the commonly used image classification benchmark datasets, which are further compared to state-of-the-art peer competitors. First of all, these benchmark datasets will be briefly described. Secondly, the peer competitors will be listed. Lastly, parameter settings of the proposed IPPSO algorithm participating in these experiments will be documented.

In this section, the benchmark datasets, peer competitors and parameter settings of the proposed IPPSO algorithm will be described.

\subsection{Benchmark Datasets}\label{sec:EPDatasets}

In these experiments, three datasets are chosen from the widely used image classification benchmark datasets to examine the performance of the proposed IPPSO method. They are the MNIST Basic (MB) \cite{DeepArchitectureEval:Larochelle}, the MNIST with Rotated Digits plus Background Images (MRDBI) \cite{DeepArchitectureEval:Larochelle} and the Convex Sets (CS) \cite{DeepArchitectureEval:Larochelle}. 

The first two benchmark datasets are two of the MNIST \cite{DocumentRecognition:LeCun} variants for classifying 10 hand-written digits (i.e., 0-9). There are a couple of reasons for using MNIST variants instead of MNIST. Firstly, as the classification accuracy of MNIST has achieved 97\%, in order to challenge the algorithm, different noises (e.g., random backgrounds, rotations) are added into these MNIST variants from the MNIST to improve the complexity of the dataset. Secondly, there are 12,000 training images and 50,000 test images in these variants, which further challenges the classification algorithms due to the much less training data but more test data. The third benchmark dataset is for recognizing the shapes of objects (i.e., convex or not), which contains 8,000 training images and 50,000 test images. Since it is a two-class classification problem comparing to 10 classes of MNIST dataset, and the images contain shapes rather than digits, it is chosen as a supplement benchmark to the two MNIST variants in order to thoroughly test the performance of the proposed IPPSO method. 

\begin{figure}[!t]
	\centering
	\includegraphics[width=2.5in]{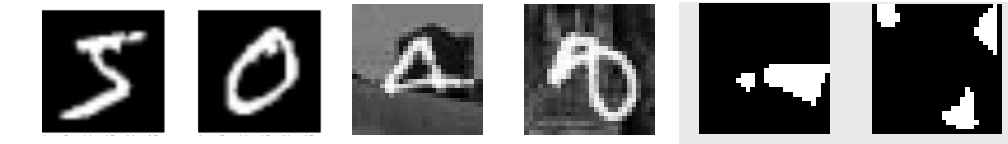}
	\caption{Examples of the three datasets. From left to right, each two images as a group are from one benchmark, and each group is from MB, MRDBI, and CS, respectively}
	\label{fig:images}
\end{figure}

Each image in these benchmarks is with the size $28 \times 28$, and examples from these three datasets are displayed in Fig. \ref{fig:images}. Another reason for choosing these three benchmark datasets is that different algorithms have reported their promising results, so it is convenient for comparing the performance of the proposed IPPSO method with these existing algorithms.

\subsection{Peer Competitors}\label{secpeer-competitors}

In the experiments, state-of-the-art algorithms, that have reported promising classification errors on the chosen benchmarks, are collected as the peer competitors of the proposed IPPSO method. To be specific, the peer competitors on the three benchmarks are CAE-2 \cite{CAE:Rifai}, TIRBM \cite{TIRBM:Sohn}, PGBM+DN1 \cite{PGBMDN1:Sohn}, ScatNet-2 \cite{ScatteringCNN:Bruna}, RandNet-2 \cite{DLBaseline:Chan}, PCANet-2 (softmax) \cite{DLBaseline:Chan}, LDANet-2 \cite{DLBaseline:Chan}, SVM+RBF \cite{DeepArchitectureEval:Larochelle}, SVM+Poly \cite{DeepArchitectureEval:Larochelle}, NNet \cite{DeepArchitectureEval:Larochelle}, SAA-3 \cite{DeepArchitectureEval:Larochelle} and DBN-3 \cite{DeepArchitectureEval:Larochelle}, which are from the literature \cite{DLBaseline:Chan} recently published and the provider of the benchmarks\footnote{\url{http://www.iro.umontreal.ca/~lisa/twiki/bin/view.cgi/Public/DeepVsShallowComparisonICML2007}}.

\subsection{Parameter Settings}

All the parameter settings are set based on the conventions in the communities of PSO \cite{PSOEPSettings:Van} and deep learning \cite{DLGuide:Hinton} which are listed in Table \ref{table:ParameterList}. 
%As both PSO and CNN are involved in the proposed IPPSO method, the parameters can be split into two parts. In regard to CNN parameters, the maximal length of CNN is set to 9 by considering the complexity of the three datasets, the maximal length of the Fully-connected layer is set to 3 based on the maximal CNN length, and the training epoch and the batch size for fitness evaluation are set to 10 and 200, respectively, in order to achieve a CNN architecture that performs well on a specific benchmark dataset in a reasonably short period. In terms of PSO parameters, 30 is used as the population size, 0.7298 is used as the inertia weight, 1.49618 is used as the coefficient of both local best and global best, and 0.1*search\_space, which is (4, 25.6) based on our IP structure design, is used as the maximal velocity for velocity clamping. 

\begin{table}[!t]
	%% increase table row spacing, adjust to taste
	\renewcommand{\arraystretch}{1.3}
	% if using array.sty, it might be a good idea to tweak the value of
	% \extrarowheight as needed to properly center the text within the cells
	\caption{Parameter list}
	\label{table:ParameterList}
	\centering
	%% Some packages, such as MDW tools, offer better commands for making tables
	%% than the plain LaTeX2e tabular which is used here.
	\begin{tabular}{|p{2.5cm}|p{3cm}|p{2cm}|}
		\hline
		Parameter Name & Parameter Meaning & Value\\
		\hline
		max\_length & maximum length of CNN layers & 9\\
		\hline
		max\_fully\_connected & maximum fully-connected layers given at least there is one fully-connected layer & 3\\
		\hline
		N & population size & 30\\
		\hline
		k & the training epoch number before evaluating the trained CNN & 10\\
		\hline
		num\_of\_batch & the batch size for evaluating the CNN & 200\\
		\hline
		c1 & acceleration coefficient array for $P_{id}$ & [1.49618,1.49618]\\
		\hline
		c2 & acceleration coefficient array for $P_{gd}$ & [1.49618,1.49618]\\
		\hline
		w & inertia weight for updating velocity & 0.7298\\
		\hline
		$v_{max}$ & maximum velocity & 4,25.6(0.1*search space)\\
		\hline
	\end{tabular}
\end{table}

The proposed IPPSO method is implemented in Tensorﬂow \cite{Tensorfow:Abadi}, and each copy of the code runs on a computer equipped with two GPU cards with the identical model number GTX1080. Due to the stochastic nature of the proposed IPPSO method, 30 independent runs are performed on each benchmark dataset, and the mean results are used for the comparisons unless otherwise specified. The experiments take two and a half hours approximately on each run of the benchmark dataset. 
%The source code will be available on https://github.com/wwwbbb8510/ipec.git for reproducing the experimental results of the proposed IPPSO method reported in this paper, and the evolved CNN architectures on each benchmark are illustrated in Section \ref{sec:EvolvedCNN}.
The evolved CNN architectures on each benchmark are illustrated in Section \ref{sec:EvolvedCNN}.

\section{Results and analysis}\label{sec:EPResults}

In this section, the classification performance along with the analysis against peer competitors, the CNN architectures learned by the proposed IPPSO method, and the related visualisation will be reported. 

%In this section, the classification performance along with the analysis against peer competitors will be shown in Section \ref{sec:Performance}, 
%the evolution time of searching the best CNN architecture and the time of training the learned CNN will be illustrated in Section \ref{sec:TrainingTime}, 
%and the CNN architectures learned by the proposed IPPSO method from the experiments for the benchmark datasets will be listed in Section \ref{sec:EvolvedCNN}. 
%the CNN architectures learned by the proposed IPPSO method from the experiments for the benchmark datasets will be listed in Section \ref{sec:EvolvedCNN}, and the classification accuracy distribution after IPPSO encoding and the PSO trajectory will be visualised in Section \ref{sec:Visulisation}. 

\subsection{Overall performance}\label{sec:Performance}

Experimental results on all the three benchmark datasets are shown in Table \ref{table:ResultComparison} where the last three rows denote the mean classification errors, the best classification errors and the standard deviations of the classification errors obtained by the proposed IPPSO method from the 30 runs, and the other rows show the best classification errors reported by peer competitors\footnote{It is a convention in the deep learning community that only the best result is reported.}. In order to conveniently investigate the comparisons, the terms “(+)” and “(-)” are provided to indicate whether the result generated by the proposed IPPSO method is better or worse than the best result obtained by the corresponding peer competitor. The term “--” means there is no available result reported from the provider or cannot be counted.

It is clearly shown in Table \ref{table:ResultComparison} that by comparing the mean classification errors of the proposed IPPSO method with the best performance of the peer competitors, IPPSO performs the second best on the MB dataset, which is only a little bit worse than LDANet-2. IPPSO is the best on the MDRBI dataset, which is the most complicated dataset among these three, and the fifth best on the CS dataset, which is not ideal but very competitive. 

\begin{table}[!t]
	%% increase table row spacing, adjust to taste
	\renewcommand{\arraystretch}{1.3}
	% if using array.sty, it might be a good idea to tweak the value of
	% \extrarowheight as needed to properly center the text within the cells
	\caption{The classification errors of the proposed IPPSO method against the peer competitors on the MB, MDRBI and CS benchmark datasets}
	\label{table:ResultComparison}
	\centering
	%% Some packages, such as MDW tools, offer better commands for making tables
	%% than the plain LaTeX2e tabular which is used here.
	\begin{tabular}{|c|c|c|c|}
		\hline
		classiﬁer & MB & MDRBI & CS\\
		\hline
		CAE-2 & 2.48(+) & 45.23(+) & --\\
		\hline
		TIRBM & -- & 35.50(+) & --\\
		\hline
		PGBM+DN-1 & -- & 36.76(+) & --\\
		\hline
		ScatNet-2 & 1.27(+) & 50.48(+) & 6.50(-)\\
		\hline
		RandNet-2 & 1.25(+) & 43.69(+) & 5.45(-)\\
		\hline
		PCANet-2 (softmax)  & 1.40(+) & 35.86(+) & 4.19(-)\\
		\hline
		LDANet-2 & 1.05(-) & 38.54(+) & 7.22(-)\\
		\hline
		SVM+RBF & 3.03(+) & 55.18(+) & 19.13(+)\\
		\hline
		SVM+Poly & 3.69(+) & 56.41(+) & 19.82(+)\\
		\hline
		NNet & 4.69(+) & 62.16(+) & 32.25(+)\\
		\hline
		SAA-3 & 3.46(+) & 51.93(+) & 18.41(+)\\
		\hline
		DBN-3  & 3.11(+) & 47.39(+) & 18.63(+)\\
		\hline
		IPPSO(mean) & 1.21 & 33 & 12.06\\
		\hline
		IPPSO(best) & 1.13 & 34.50 & 8.48\\
		\hline
		IPPSO(standard deviation) & 0.103 & 2.96 & 2.25\\
		\hline
	\end{tabular}
\end{table}

%\subsection{Training time}\label{sec:TrainingTime}
%
%In the experiments, we evolve the architectures of CNNs first and then train the learned architecture in order to achieve the best test accuracy, so the total training time of one run is comprised of the evolution time and the CNN training time. For all the three benchmark datasets, the evolution time is almost two hours - 114 minutes, 114 minutes and 118 minutes for the MB, MDRBI and CS benchmark, respectively, in one of the 30 runs; in terms of the CNN training time, we train each of the learned architectures of CNNs 100 epochs, and it nearly takes half an hour - 28 minutes, 27 minutes and 24 minutes for the MB, MDRBI and CS benchmark, respectively, in one of the 30 runs. 

\subsection{Evolved CNN Architectures}\label{sec:EvolvedCNN}

Although the proposed IPPSO method is performed on each benchmark with 30 independent runs, only one is chosen on each benchmark for this description purpose shown from Table \ref{table:EvolvedMBCNN} to Table \ref{table:EvolvedConvexCNN}. Since Disabled layers have been removed during the decoding process, they do not show up in the learned CNN architectures. Therefore, it turns out that IPPSO is able to learn a variable-length CNN architecture, which can be obviously seen from the listed architectures - 6 CNN layers for the MB and CS benchmark and 8 CNN layers for the MDRBI benchmark. 

\begin{table}[!t]
	%% increase table row spacing, adjust to taste
	\renewcommand{\arraystretch}{1.3}
	% if using array.sty, it might be a good idea to tweak the value of
	% \extrarowheight as needed to properly center the text within the cells
	\caption{An evolved architecture for the MB benchmark}
	\label{table:EvolvedMBCNN}
	\centering
	%% Some packages, such as MDW tools, offer better commands for making tables
	%% than the plain LaTeX2e tabular which is used here.
	\begin{tabular}{|c|c|}
		\hline
		Layer type & Configuration\\
		\hline
		conv & Filter size: 2, Stride size: 1, feature maps: 26\\
		\hline
		conv & Filter size: 6, Stride size: 3, feature maps: 82\\
		\hline
		conv & Filter size: 8, Stride size: 4, feature maps: 114\\
		\hline
		conv & Filter size: 7, Stride size: 4, feature maps: 107\\
		\hline
		full & Neurons: 1686\\
		\hline
		full & Neurons: 10\\
		\hline
	\end{tabular}
\end{table}

\begin{table}[!t]
	%% increase table row spacing, adjust to taste
	\renewcommand{\arraystretch}{1.3}
	% if using array.sty, it might be a good idea to tweak the value of
	% \extrarowheight as needed to properly center the text within the cells
	\caption{An evolved architecture for the MDRBI benchmark}
	\label{table:EvolvedMDRBICNN}
	\centering
	%% Some packages, such as MDW tools, offer better commands for making tables
	%% than the plain LaTeX2e tabular which is used here.
	\begin{tabular}{|c|c|}
		\hline
		Layer type & Configuration\\
		\hline
		conv & Filter size: 2, Stride size: 1, feature maps: 32\\
		\hline
		conv & Filter size: 6, Stride size: 3, feature maps: 90\\
		\hline
		conv & Filter size: 7, Stride size: 4, feature maps: 101\\
		\hline
		conv & Filter size: 7, Stride size: 4, feature maps: 97\\
		\hline
		pool & Kernel size: 4, Stride size: 4, Type: Average\\
		\hline
		conv & Filter size: 5, Stride size: 3, feature maps: 68\\
		\hline
		full & Neurons: 1577\\
		\hline
		full & Neurons: 10\\
		\hline
	\end{tabular}
\end{table}

\begin{table}[!t]
	%% increase table row spacing, adjust to taste
	\renewcommand{\arraystretch}{1.3}
	% if using array.sty, it might be a good idea to tweak the value of
	% \extrarowheight as needed to properly center the text within the cells
	\caption{An evolved architecture for the CS benchmark}
	\label{table:EvolvedConvexCNN}
	\centering
	%% Some packages, such as MDW tools, offer better commands for making tables
	%% than the plain LaTeX2e tabular which is used here.
	\begin{tabular}{|c|c|}
		\hline
		Layer type & Configuration\\
		\hline
		conv & Filter size: 1, Stride size: 1, feature maps: 11\\
		\hline
		conv & Filter size:7, Stride size: 4, feature maps: 108\\
		\hline
		conv & Filter size: 1, Stride size: 1, feature maps: 8\\
		\hline
		conv & Filter size: 6, Stride size: 3, feature maps: 92\\
		\hline
		full & Neurons: 906\\
		\hline
		full & Neurons: 2\\
		\hline
	\end{tabular}
\end{table}

\subsection{Visualisation}\label{sec:Visulisation}

In order to achieve a better understanding of the proposed IPPSO method, we visualise two parts of the evolutionary process - the accuracy distribution of the PSO vectors, where the architectures of CNNs are encoded, and the PSO trajectory of the evolving process.

\begin{figure}[!htp]
	\centering
	\subfloat[Surface]{\includegraphics[width=0.19\textwidth]{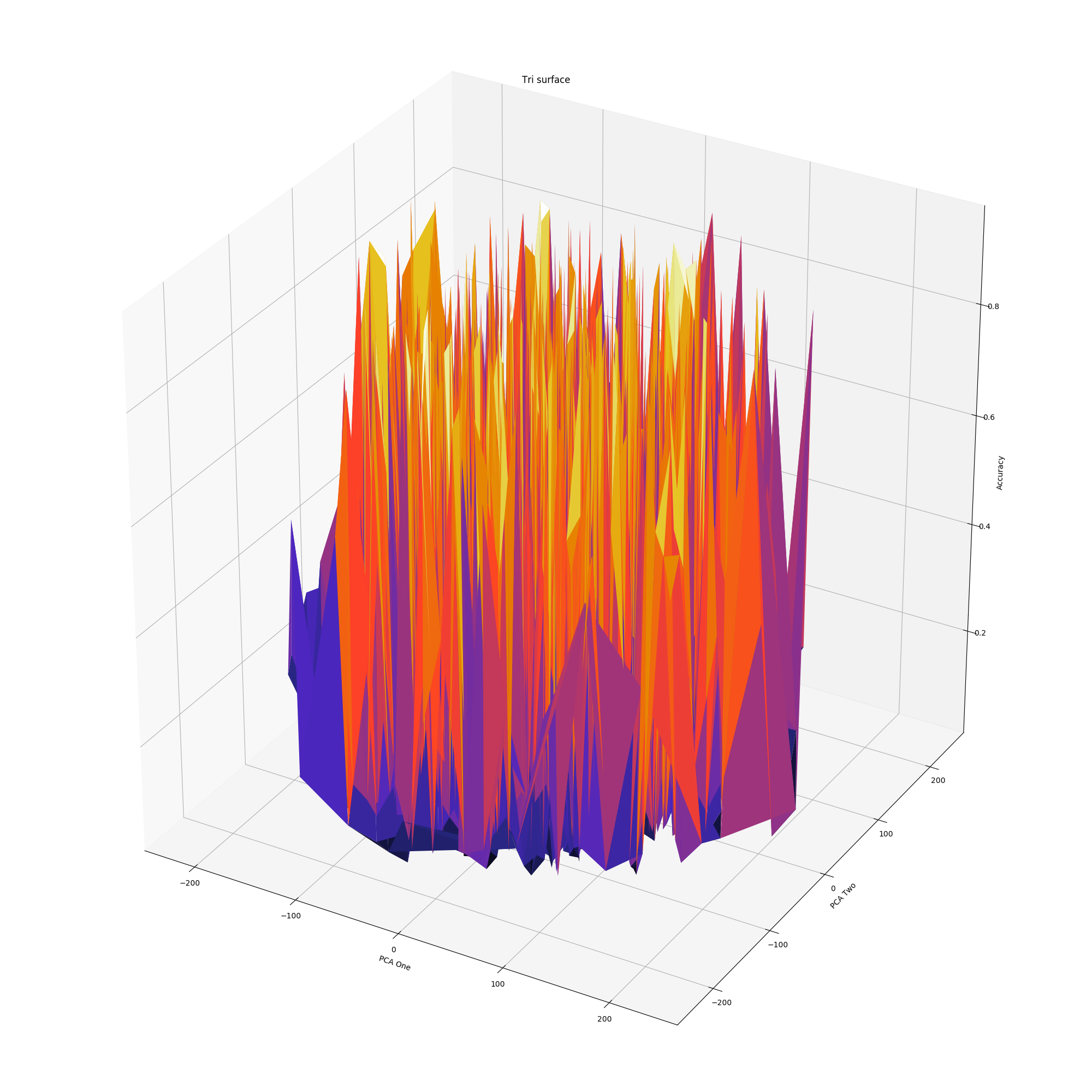}%
		\label{fig:Visual3DSurface}}
	\hfil
	\subfloat[Trajectory]{\includegraphics[width=0.19\textwidth]{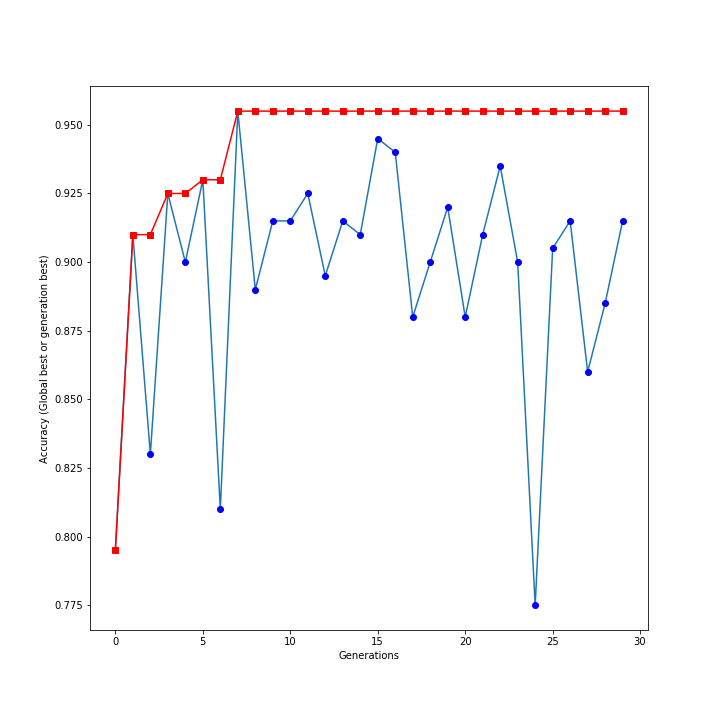}%
		\label{fig:VisualTrajectory}}
	
	\caption{(\ref{fig:Visual3DSurface}): The surface of CNN accuracies after training 10 epochs with IPPSO encoding; (\ref{fig:VisualTrajectory}): PSO Trajectory.}
\end{figure}

In terms of the accuracy distribution, first of all, we obtained the PSO vectors and their corresponding accuracies from 10 runs of the experiments; in addition, the first two principal components from Principal Component Analysis (PCA) are extracted for the usage of visualisation. A 3-D triangulated surface with the data containing the first two components and the corresponding accuracy is plotted shown in Fig. \ref{fig:Visual3DSurface}. It is observed that there are a lot of steep hills on the surface whose summits are at the similar level, so it means that there are quite a number of local optima, but most of them are very close to each other, which means that those local optima are acceptable as a good solution of the task. 

%\begin{figure}[!t]
%	\centering
%	\includegraphics[width=3in]{ipec_pca_plot_2d_surface}
%	\caption{The surface of CNN accuracies after training 10 epochs with IPPSO encoding}
%	\label{fig:Visual3DSurface}
%\end{figure}

Regarding the trajectory, the best result of the particles of each generation and the global best in each generation from one run of the experiments are obtained and plotted in blue colour and red colour, respectively, in Fig. \ref{fig:VisualTrajectory}. It can be seen that after only a few generations, the global best is found, after which the particles are still flying in the search space, but none of them can obtain a better accuracy, which means the optimum has been reached by PSO after only a few steps. Even though the surface of the optimisation task shown in Fig. \ref{fig:Visual3DSurface} is extremely complicated, the PSO method with only 30 particles can climb up to the optimum very quickly, which proves the effectiveness and efficiency of PSO on optimisation tasks.

\section{Conclusions}\label{sec:Conclusion}

The goal of this paper was to develop a new PSO approach with variable length to automatically evolve the architectures of CNNs for image classification problems. This goal has been successfully achieved by proposing a new encoding scheme of using a network interface containing an IP address and its corresponding subnet to carry the configurations of a CNN layers, the design of four subnets including a disabled subnet in order to simulate a variable-length PSO, and an efficient fitness evaluation method by using partial dataset. This approach was examined and compared with 12 peer competitors including the most state-of-the-art algorithms on three benchmark datasets commonly used in deep learning and the experimental results show that the proposed IPPSO method can achieve a very competitive accuracy by outperforming all others on the MDRBI benchmark dataset, being the second-best on the MNIST benchmark dataset and ranking above the middle line on the CS benchmark dataset.

The most important improvement from the traditional PSO to the novel IPPSO proposed in the paper is to invent the new encoding strategy of using network interface. Since the subnet in the interface can distinguish any type of layers, any layer configurations can be encoded into the IP address and the length of the IP address can be easily extended to 4 bytes (the length of real IP addresses) or even more, the IPPSO method has the ability of encoding any type of Deep Neural Network layers. In addition, the particle length of the IPPSO method can be easily made variable by simply introducing a disabled layer which could be deemed as another major improvement as it breaks the obstacle of traditional PSO being fix-length.

In this paper, we have investigated the proposed IPPSO method for evolving deep CNN and it is proved of obtaining promising results. Based on this research, there are a couple of further researches that are worth doing. Firstly, as this paper is mainly to propose the novel encoding strategy of the IPPSO, it will be interesting to see how different PSO topologies\footnote{Fully-connected topology is used in this paper as it is easy to be implemented.} will affect the performance of the IPPSO and design the best topology for it. Secondly, we will also investigate how the proposed IPPSO algorithm performs for evolving recurrent neural networks, which are powerful tools for addressing sequential data tasks, such as language processing problems.


%\section*{Acknowledgement}

%The authors would like to thank all the supervisors. 



%\bibliographystyle{IEEEtran}

\end{document}